\documentclass[draftclasofoot, onecolumn]{IEEEtran}
\IEEEoverridecommandlockouts

\usepackage[noadjust]{cite}
\usepackage[bookmarks=true,breaklinks=true,colorlinks,linkcolor=red,citecolor=blue,urlcolor=BlueViolet]{hyperref}
\usepackage{multirow}
\usepackage{amsmath,amssymb,amsfonts}
\usepackage[ruled,vlined]{algorithm2e}
\newcommand{\mycommentstyle}[1]{\color[HTML]{0671b9}{#1}}
\SetKwComment{Comment}{\mycommentstyle{// }}{}
\usepackage{graphicx}
\usepackage{threeparttable}
\usepackage{subfigure}
\usepackage{textcomp}
\usepackage{xcolor}
\def\BibTeX{{\rm B\kern-.05em{\sc i\kern-.025em b}\kern-.08em
    T\kern-.1667em\lower.7ex\hbox{E}\kern-.125emX}}
\usepackage{booktabs}
\usepackage{tablefootnote}
\usepackage{blindtext}
\usepackage{float}
\usepackage{algorithmic}
\usepackage{textcomp}
\usepackage{booktabs}

\begin{document}

\title{A Nonlinear Low-rank Representation Model with Convolutional Neural Network for Imputing Water Quality Data\footnote{This paper is a preprint of a paper submitted to 9th International Conference on Electronic Information Technology and Computer Engineering (EITCE 2025), and is subject to Institution of Engineering and Technology Copyright. If accepted, the copy of record will be available at IET Digital Library.}}

\author{\IEEEauthorblockN{Xin Liao}\\
	\IEEEauthorblockA{\textit{College of Computer and Information Science}
		\textit{Southwest University}
		Chongqing, China \\
		lxchat26@gmail.com}
\\
	\IEEEauthorblockN{Bing Yang}\\
	\IEEEauthorblockA{\textit{Chongqing Eco-Environment Monitoring Center}
		Chongqing, China \\
		cq\_yangbing@163.com}
\\
	\IEEEauthorblockN{Cai Yu*}\\
	\IEEEauthorblockA{\textit{Chongqing Eco-Environment Monitoring Center}
		Chongqing, China \\
		cyscut@foxmail.com}
}

\maketitle

\begin{abstract}
The integrity of Water Quality Data (WQD) is critical in environmental monitoring for scientific decision-making and ecological protection. However, water quality monitoring systems are often challenged by large amounts of missing data due to unavoidable problems such as sensor failures and communication delays, which further lead to water quality data becoming High-Dimensional and Sparse (HDS). Traditional data imputation methods are difficult to depict the potential dynamics and fail to capture the deep data features, resulting in unsatisfactory imputation performance. To effectively address the above issues, this paper proposes a Nonlinear Low-rank Representation model (NLR) with Convolutional Neural Networks (CNN) for imputing missing WQD, which utilizes CNNs to implement two ideas: a) fusing temporal features to model the temporal dependence of data between time slots, and b) Extracting nonlinear interactions and local patterns to mine higher-order relationships features and achieve deep fusion of multidimensional information. Experimental studies on three real water quality datasets demonstrate that the proposed model significantly outperforms existing state-of-the-art data imputation models in terms of estimation accuracy. It provides an effective approach for handling water quality monitoring data in complex dynamic environments.
\end{abstract}

\begin{IEEEkeywords}
	Water quality data, High-dimensional and sparse, Data imputation, low-rank representation model, Causal convolutional, hyperparameters adaptation.
\end{IEEEkeywords}

\section{Introduction}
In the field of environmental monitoring, the integrity of Water Quality Data (WQD) is crucial for scientific decision-making and ecological protection~\cite{zeng2016cost}. By collecting WQD regularly, water quality trends are effectively identified and timely measures are taken to ensure water quality. However, water quality monitoring systems in practice frequently face the problem of missing data due to various factors such as sensor failures, aging equipment, and communication delays. This problem has become more prominent with the increasing number of sensor deployments and the High-Dimensional and Sparse (HDS)~\cite{r1,r2,r3,r4,r5,r6, r57, r94,r95,r96,r97,r98,r99,r100} nature of monitoring data. Missing data not only affects the accuracy and reliability of the water quality monitoring system but also makes the subsequent data analysis and decision-making process complex and difficult. Therefore, how to accurately impute the missing values in the HDS WQD has become a thorny task~\cite{r21,r8,r9,r60,r61,r62}.

In recent years, researchers have proposed a variety of missing data imputation models to effectively fill in the missing values in the WQD, Murti~\textit{et al.}~\cite{8987530} use K-Nearest Neighbours (KNN) imputation to capture the complex relationships in the data, but accurately identifying the true neighbours becomes difficult when dealing with HDS data, which results in the accuracy of the imputation results being affected. Rahman \textit{et al.}~\cite{rahman2013missing} utilize a decision tree to model the inter-feature of complex interaction relationships to improve the imputation accuracy. However, this method is prone to the risk of overfitting. Brand \textit{et al.}~\cite{brand2021multi} portray the nonlinear relationships in the WQD data by Support Vector Machines (SVM) imputation and improve the robustness by regularizing the parameters, but for large-scale datasets, the method is challenging in terms of computational complexity~\cite{r14,r53,r16,r30,r70,r71,r72,r73,r74,r75,r76}.

\begin{figure*}[t]
	\centering
	\includegraphics[width=0.8\linewidth]{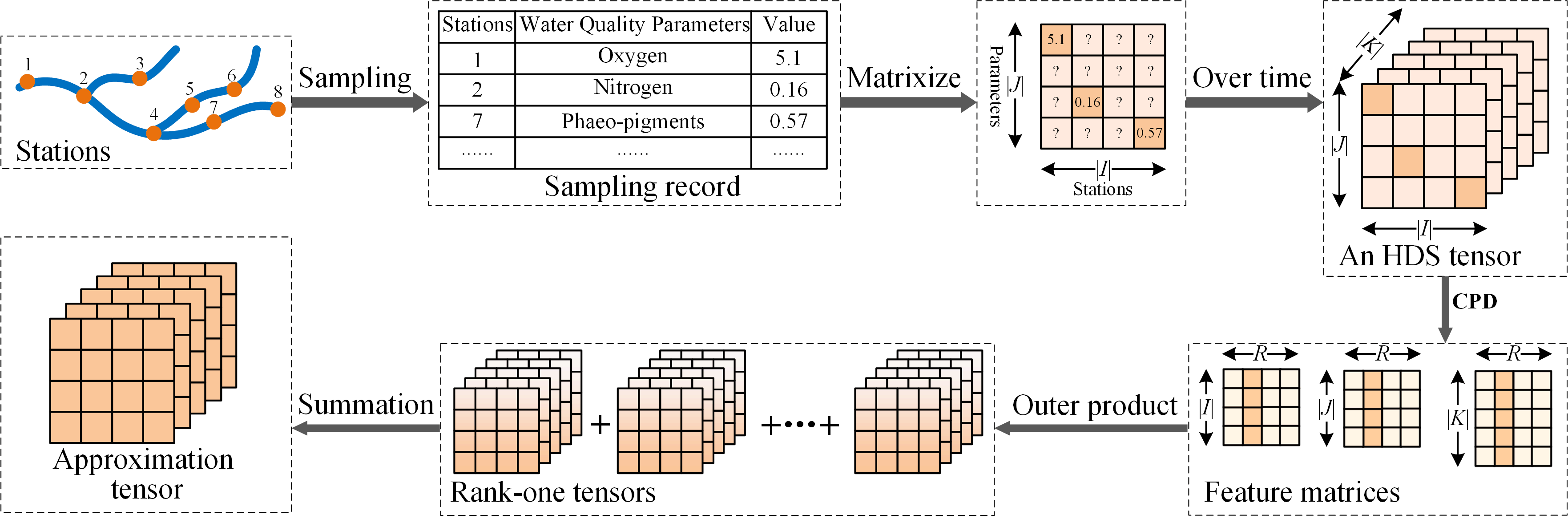}
	\caption{An instance of a TLR model representing the HDS tensor constructed from WQD.}
	\label{wqd}
\end{figure*}

Previous studies have shown that the Tensor Low-rank Representation (TLR) model~\cite{r22, r10, r12, r65} is highly effective in handling high-dimensional sparse data by providing compact vector representations of data features in a low-rank latent space. By exploiting the underlying low-rank structure of multi-dimensional data, TLR enables efficient extraction of discriminative information and facilitates accurate recovery of missing values. On the other hand, WQD can be naturally modeled as HDS tensors due to its multi-source heterogeneity, temporal continuity, and missing values. Moreover, the convolutional neural network (CNN)~\cite{r20, r46, r52} is particularly effective in modeling sequential dependencies, making it well-suited for temporal data owing to its local receptive fields and shared weights~\cite{r17, r18, r19, r24, r31, r32,r101,r102,r103,r104,r106,r107,r40}.
Based on the above observation, this paper proposes a novel Nonlinear Low-rank Representation model with CNN (CLR-CNN). Overall, the main contributions are as follows:
\begin{itemize}
	\item Integrating time-series features through convolutional operations to model the temporal dependencies between different time slices of water quality data, thereby more accurately recovering missing values and capturing potential dynamic trends;
	\item Leveraging the advantages of convolutional networks in extracting local patterns and nonlinear interactions, the method uncovers high-order associative features between different variables, enabling deep fusion of multi-dimensional information and significantly improving the model’s ability to express complex data structures and its imputation accuracy; and.
	\item An experimental study on three datasets. It demonstrates that the proposed NLR-CNN model significantly outperforms several state-of-the-art imputation models in terms of estimation accuracy for missing values in the WQD.
\end{itemize}

Section~\ref{back} introduces the background, Section~\ref{model} describes the proposed NLR-CNN model, Section~\ref{comparisons} provides the experimental study, and Section~\ref{conclusions} concludes this paper.

\section{Background}\label{back}
\subsection{Notation system}

\begin{table}[t]
	\caption{The definition of the notation}\label{t1}
	\centering
	\begin{tabular}{@{}cl@{}}
		\toprule
		Notation & Description \\ \midrule
		$ I $ & Set of sampling stations. \\
		$ J $ & Set of WQD parameters. \\
		$ K $ & A set of time slots. \\
		$\textbf{X} $ & The HDS tensor. \\
		$\textbf{Z} $ & The low-rank approximation tensor.  \\ 
		$\textbf{A} $ & Rank-one tensors. \\
		$ \mathrm{S}, \mathrm{U}, \mathrm{V} $ & Feature matrices for $ I, J, K $. \\
		$ \circ $ & Outer product. \\
	 \bottomrule
	\end{tabular}
	\label{symbol}
\end{table}

The notation adopted in this paper is listed in Table~\ref{t1}. Specifically, tensors are represented by bold uppercase letters, matrices are represented by standard uppercase letters, vectors are represented by bold italic lowercase letters, and scalars are represented by italic lowercase letters

\subsection{Problem formulation}

Fig.~\ref{wqd} presents the flow of data imputation of a TRL model for an HDS tensor constituted by the WQD, where the HDS tensor is defined as:

\textbf{Definition 1.} (HDS tensor of the WQD). The water quality parameters of the sampling stations are recorded via the sampling stations. By constructing the matrix of sampling records for each time slot and arranging them over time, a tensor $ \textbf{X} \in \mathbb{R}^{|I|\times|J|\times|K|} $ is constructed. Let $ \Omega $ and $ \Upsilon $ be known and unknown data sets. The tensor is an HDS tensor if $ |\Upsilon|  \gg   |\Omega| $, where an element $ x_{ijk} $ denotes the quantized value of the index $ (i,j,k) $.

By the Canonical Polyadic Decomposition (CPD)~\cite{r36,r37,r38,r39,r28} principle, the HDS tensor $ \textbf{X} $ is decomposed into three low-rank feature matrices, $ {\rm{S}} \in \mathbb{R}^{|I| \times R} $, $ {\rm{U}} \in \mathbb{R}^{|J| \times R} $, and $ {\rm{V}} \in \mathbb{R}^{|K| \times R} $, to represent the three modes. To get the $ r $-th rank-one tensor $ \textbf{A}_{r} $, the $ r $-th column vector of the three feature matrices performs the outer product~\cite{r25,r26,r29,r33} as follows:
\begin{equation}\label{e1}
	\tilde{\textbf{A}}_r = \textbf{s}_{:r} \circ \textbf{u}_{:r} \textbf{v}_{:r}.
\end{equation}
Further, the low-rank approximation tensor \textbf{Z} is constructed~\cite{r41,r42,r43,r44,r45} as below:
\begin{equation}
	\textbf{Z} = \sum\limits_{r = 1}^R {\textbf{\textit{s}}_{:r} \circ \textbf{\textit{u}}_{:r} \circ  \textbf{\textit{v}}_{:r}}.
\end{equation}

To measure the discrepancy between the original HDS tensor and the approximation tensor, the Euclidean distance~\cite{r48, r54,r55} is typically employed, and the objective function is constructed as follows:
\begin{equation}
	\varepsilon  =  ||{\textbf{X} - \textbf{Z}}||_F^2.
\end{equation}

By defining the objective function on a known set of elements, it is reformulated as:
\begin{equation}
	\varepsilon  = \frac{1}{2}{\sum\limits_{{x_{ijk}} \in \Lambda } {\left( {{x_{ijk}} - \sum\limits_{r = 1}^R {{s_{ir}}{u_{jr}}{v_{kr}}} } \right)} ^2}.
\end{equation}

\begin{figure*}[t]
	\centering
	\includegraphics[width=0.8\linewidth]{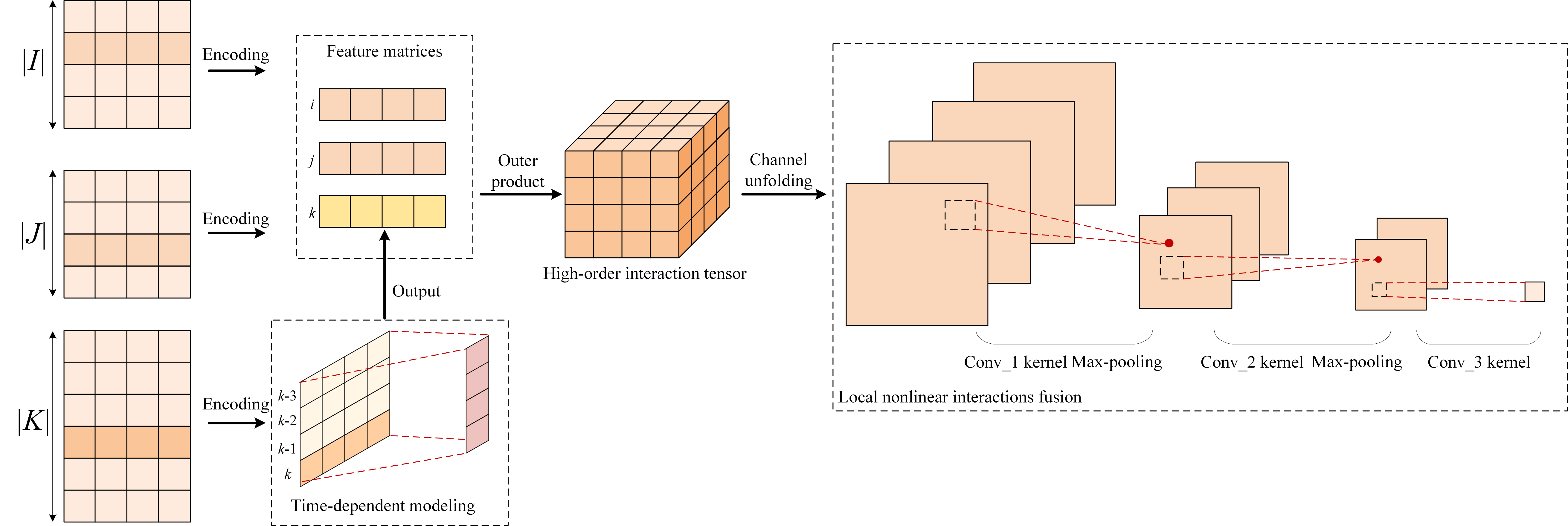}
	\caption{The overall framework of NLR-CNN.}
	\label{nlr}
\end{figure*}

\section{Methodology}\label{model}
Fig.~\ref{nlr} illustrates our proposed CLR-CNN model, which is a higher-order interaction learning framework that fuses time-dependent modeling with local nonlinear interaction fusion. Specifically, it first constructs three feature matrices based on the TLR model to represent the sampling station features, water quality parameter features, and temporal features, which serve as inputs to the CLR-CNN model. Then, it models the temporal dependence of the temporal features to effectively mine the dynamic dependence of entity interactions over time and provide rich temporal context information. Finally, the outer product operation is adopted to construct a 3D higher-order interaction tensor to explicitly express the higher-order combinatorial relationships between Entities. Through channel unfolding, the higher-order interaction tensor is fed into the CNN module to further model local nonlinear interactions, and a multilayer convolutional network is introduced to capture local nonlinear patterns and boost the representation ability of complex structural information The overall approach achieves time-dependent modeling and spatial structure fusion, which enhances the model’s ability to express and generalize multi-dimensional interaction relations.

\subsection{Time-dependent modeling}

For instance, $ x_{ijk} $ in the water quality dataset, its temporal feature vector encodes the temporal feature matrix to obtain $\textbf{\textit{v}}_k$. In general, WQD are significantly time-dependent and time-series dynamic~\cite{r86}, and observations at different time slots are often strongly correlated~\cite{r56,r58,r59,r77,r78,r79}. It is difficult to adequately capture the evolutionary trends and potential patterns of water quality indicators by only considering the information of a single time slot in the modeling process. Therefore, we introduce a time-dependent modeling mechanism, which encodes and integrates water quality observations from multiple time slots in history to construct the representation with temporal contextual features. Specifically, we encode the current time slot $ k $ and the previous $ t $ time slot and derive a semantically rich feature representation via a convolution operation as:
\begin{equation}
	{\tilde v_k} = {\mathop{\rm Re}\nolimits} LU\left( {w * \left[ {{v_{k - t + 1}};{v_{k - t + 2}};...;{v_k}} \right]} \right),
\end{equation}
where $ * $ denotes the convolutional operator and $ {\textbf{\textit{w}}_r} \in \mathbb{R}^{C} $ denotes the weight vector in the $ r $-th channel. For an element $ \bar{v}_{kr} $ in the feature vector, it is fine-grained formulated as:
where $ \tilde{v}_k $ denotes a feature representation incorporating temporal information, ReLU denotes an activation, $ * $ denotes a convolution operator, and $ \textbf{\underline{w}} $ denotes a weight vector. 

\subsection{Local nonlinear iteration fusion}
To further model the complex, high-order dependencies among different entities (e.g., monitoring stations, indicators, and time), we design a local nonlinear interaction fusion module to capture fine-grained and localized semantic patterns. After obtaining the temporal-aware feature representation, we perform an outer product operation among the encoded embedding of the involved entities to construct a high-order interaction tensor $ \textbf{H} $ with the dimension $ R \times R \times R $, where each mode corresponds to a specific dimension. To effectively extract local and nonlinear dependencies from this high-dimensional tensor, we apply a 2D CNN across different temporal features is treated as a semantic interaction map that reveals latent correlations across dimensions. Formally, the convolution process is formulated as follows:
\begin{equation}
	{z_{ijk}} = Sigmoid\left( {Conv\_1\left( {Conv\_2\left( {Conv\_3\left( H \right)} \right)} \right)} \right),
\end{equation}

where Sigmoid denotes the activation function~\cite{r63,r64}, Conv\_1 and Conv\_2 are the 2D convolutional operations with kernel size 3$ \times $3, Conv\_1 is the 2D convolutional operation with kernel size 2$ \times $2. Note that we impose a 2×2 max-pooling operation after the 3$ \times $3 convolution operator to further compress the spatial dimensions of the feature map, enhance the local feature aggregation capability, and effectively reduce the computational complexity~\cite{r80,r81,r82,r83,r84,r85, r87,r88,r89,r90,r91,r92,r93}.

\section{Experimental results}\label{comparisons}
\subsection{Datasets}
In this study, we conduct comprehensive evaluations of the proposed model using three real-world water quality datasets Specifically, D1 includes 129406 surface-level observations, D2 comprises 121218 measurements at middle depth, and D3 consists of 129415 bottom-level records. Each dataset is partitioned into training $ (\Gamma) $, validation $ (\Lambda) $, and the testing $ (\Phi) $ sets following a 1:2:7 ratio.

\subsection{Evaluation metrics}
Root Mean Squared Error (RMSE) and Mean Absolute Error (MAE)~\cite{r66,r67,r68,r69} are adopted as metrics, and they are given as follows:
\begin{equation}
	\begin{array}{l}
		{\rm{RMSE}} = \sqrt {{{\sum\limits_{{x_{ijk}} \in \Phi } {{{\left( {{x_{ijk}} - {{z}_{ijk}}} \right)}^2}} } \mathord{\left/
					{\vphantom {{\sum\limits_{{x_{ijk}} \in \Phi } {{{\left( {{x_{ijk}} - {{\tilde x}_{ijk}}} \right)}^2}} } {\left| \Phi  \right|}}} \right.
					\kern-\nulldelimiterspace} {\left| \Phi  \right|}}} ;\\
		{\rm{MAE}} = {{\sum\limits_{{x_{ijk}} \in \Phi } {{{\left| {{x_{ijk}} - {{z}_{ijk}}} \right|}_{{\rm{abs}}}}} } \mathord{\left/
				{\vphantom {{\sum\limits_{{x_{ijk}} \in \Phi } {{{\left| {{x_{ijk}} - {{\tilde x}_{ijk}}} \right|}_{{\rm{abs}}}}} } {\left| \Phi  \right|}}} \right.
				\kern-\nulldelimiterspace} {\left| \Phi  \right|}},
	\end{array}
\end{equation}
where $ |\cdot|_{\rm abs} $ denotes the absolute value.

\subsection{Results}

\begin{table}[t]
	\caption{The imputation accuracy for five tested models}
	\begin{center}
		\begin{tabular}{ccccccc}
			\hline
			\textbf{Dataset} &  & \textbf{M1} & \textbf{M2} & \textbf{M3} & \textbf{M4} & \textbf{M5} \\ \hline
			\multirow{2}{*}{\textbf{D1}} & \textbf{RMSE} & \textbf{0.0203} & 0.0256 & 0.0260 & 0.0273 & 0.0267 \\
			& \textbf{MAE}  & \textbf{0.0135} & 0.0162 & 0.0148 & 0.0162 & 0.0163 \\
			\multirow{2}{*}{\textbf{D2}} & \textbf{RMSE} & \textbf{0.0233} & 0.0262 & 0.0266 & 0.0287 & 0.0270 \\
			& \textbf{MAE}  & \textbf{0.0142} & 0.0175 & 0.0161 & 0.0177 & 0.0176 \\
			\multirow{2}{*}{\textbf{D3}} & \textbf{RMSE} & \textbf{0.0244} & 0.0258 & 0.0262 & 0.0277 & 0.0259 \\
			& \textbf{MAE}  & \textbf{0.0151} & 0.0176 & 0.0159 & 0.0175 & 0.0166 \\\hline
		\end{tabular}
		\label{tp}
	\end{center}
\end{table}

We evaluate the proposed NLR-CNN model against four state-of-the-art imputation approaches across three datasets in imputation accuracy. The comparison includes: a) M1: our proposed NLR-CNN model; b) M2: a biased low-rank representation-based method~\cite{r6} that utilizes a nonnegative update algorithm for parameter learning; c) M3: a robust approach~\cite{ye2021outlier} leveraging the Cauchy loss function to handle noise and outliers; d) M4: a multilinear imputation technique~\cite{wang2016multi} incorporating an integrated reconfiguration optimization strategy; and e) M5: a multidimensional method M5~\cite{su2021tensor} based on the alternating least squares optimization algorithm. All models are configured with a rank of 10 and a maximum of 1000 training iterations. Early stopping is applied when the difference in error between two successive iterations falls below $ 10^{-5} $. The statistical outcomes are summarized in Table~\ref{tp}. From these results, we can find that the M1 model consistently exhibits superior imputation accuracy when compared with the other evaluated methods. For example, the RMSE and MAE of M1 on D1 are 0.0203 and 0.0135, respectively. These results represent improvements of 20.70\% and 16.67\%, 21.92\% and 8.78\%, 25.64\% and 16.67\%, and 23.97\% and 17.18\% over M2-5, whose RMSE and MAE values are 0.0256 and 0.0162, 0.0260 and 0.0148, 0.0273 and 0.0162, and 0.0267 and 0.0163, respectively. Similarly, M1 achieves the best performance on D2, with an RMSE of 0.0233 and MAE of 0.0142. Compared to M2-5, the improvements are 11.07\% and 18.86\%, 12.41\% and 11.80\%, 18.81\% and 19.77\%, and 13.70\% and 19.32\%, respectively. These consistent gains highlight the robustness and effectiveness of the proposed NLR-CNN model in handling missing data across different water quality layers.

The analysis indicates that the NLR-CNN model outperforms several imputation models in estimation accuracy.

\section{Conclusions}\label{conclusions}
In this study, we proposed a novel Nonlinear Low-rank Representation model with Convolutional Neural Networks (NLR-CNN) to address the challenges of imputing missing values in HDS water quality monitoring data. By integrating temporal feature modeling and nonlinear interaction extraction, the model effectively captures both time dependencies and complex local patterns in the data. The experiments conducted on three real-world datasets validate the superior performance of the proposed method over the existing state-of-the-art imputation model. These results highlight the potential of our approach for improving the reliability and completeness of water quality datasets~\cite{r11}.

\bibliographystyle{IEEEtran}
\bibliography{CLR}

\vspace{12pt}

\end{document}